\documentclass[conference]{IEEEtran}
\IEEEoverridecommandlockouts
\usepackage{cite}
\usepackage{amsmath,amssymb,amsfonts}
\usepackage{graphicx}
\usepackage{textcomp}
\usepackage{xcolor}
\usepackage{algorithm}
\usepackage{multirow}
\usepackage{multicol}
\usepackage{subfigure}
\usepackage{amssymb}
\usepackage{bbding}
\usepackage{pifont}
\usepackage{wasysym}
\usepackage{booktabs}
\usepackage{etoolbox}
\usepackage{algorithm}
\usepackage{algpseudocode}
\usepackage{graphics}
\usepackage{epsfig}
\usepackage{amsthm}
\usepackage[misc]{ifsym}
\makeatletter
\patchcmd{\@makecaption}
  {\scshape}
  {}
  {}
  {}
\makeatletter
\patchcmd{\@makecaption}
  {\\}
  {.\ }
  {}
  {}
\makeatother

\newcommand{\our}{\textsc{Subg-Con}}
\theoremstyle{definition}
\newtheorem{myDef}{Definition}

\def\BibTeX{{\rm B\kern-.05em{\sc i\kern-.025em b}\kern-.08em
    T\kern-.1667em\lower.7ex\hbox{E}\kern-.125emX}}
\begin{document}

\title{Sub-graph Contrast for Scalable Self-Supervised Graph Representation Learning\\
}

\author{
\IEEEauthorblockN{Yizhu Jiao\textsuperscript{1}, Yun Xiong\textsuperscript{1, 2}\textsuperscript{(\Letter)} , Jiawei Zhang\textsuperscript{3}, Yao Zhang\textsuperscript{1}, Tianqi Zhang\textsuperscript{1}, Yangyong Zhu\textsuperscript{1, 2} }
\IEEEauthorblockA{
\textit{\textsuperscript{1}Shanghai Key Laboratory of Data Science, School of Computer Science, Fudan University, China} \\
\textit{\textsuperscript{2}Shanghai Institute for Advanced Communication and Data Science, Fudan University, China} \\
\textit{\textsuperscript{3}IFM Lab, Department of Computer Science, Florida State University, FL, USA} \\}
{Email: \textsuperscript{1}\{yzjiao18, yunx, yaozhang18, tqzhang18, yyzhu\}@fudan.edu.cn, \textsuperscript{3}jiawei@ifmlab.org} \\   
\thanks{\textsuperscript{\Letter} Corresponding author.}
}


\maketitle

\begin{abstract}

Graph representation learning has attracted lots of attention recently. Existing graph neural networks fed with the complete graph data are not scalable due to limited computation and memory costs. Thus, it remains a great challenge to capture rich information in large-scale graph data. Besides, these methods mainly focus on supervised learning and highly depend on node label information, which is expensive to obtain in the real world. As to unsupervised network embedding approaches, they overemphasize node proximity instead, whose learned representations can hardly be used in downstream application tasks directly. In recent years, emerging self-supervised learning provides a potential solution to address the aforementioned problems. However, existing self-supervised works also operate on the complete graph data and are biased to fit either global or very local (1-hop neighborhood) graph structures in defining the mutual information based loss terms.

In this paper, a novel self-supervised representation learning method via \underline{Sub-g}raph \underline{Con}trast, namely {\our}, is proposed by utilizing the strong correlation between central nodes and their sampled subgraphs to capture regional structure information. Instead of learning on the complete input graph data, with a novel data augmentation strategy, {\our} learns node representations through a contrastive loss defined based on subgraphs sampled from the original graph instead. Compared with existing graph representation learning approaches, {\our} has prominent performance advantages in weaker supervision requirements, model learning scalability, and parallelization. Extensive experiments verify both the effectiveness and the efficiency of our work compared with both classic and state-of-the-art graph representation learning approaches on multiple real-world large-scale benchmark datasets from different domains. 
\end{abstract}


\section{Introduction}

    Graph representation learning \cite{hamilton2017representation} has attracted much attention recently. Its basic idea is to extract the high-dimensional information in graph-structured data and embed it into low-dimensional vector representations. These node representation vectors can be potentially used in various downstream tasks such as node classification \cite{kipf2016semi}, link prediction \cite{grover2016node2vec}, graph classification \cite{lee2019self}, and graph alignment \cite{jiao2019collective}. Graph representation learning problems have been studied on graph data from many different domains such as social networks \cite{chen2018fastgcn}, chemical molecular graphs \cite{liao2019lanczosnet}, and bio-medical brain graphs \cite{wang2017structural}.

    Most existing successful methods are based on graph neural networks (GNNs) \cite{kipf2016semi, DBLP:journals/corr/abs-1710-10903, wu2019simplifying, qu2019gmnn} , which learn nodes' contextualized representations via effective neighborhood information aggregation. These methods usually take a complete graph as the input, which can hardly be applied to large-scale graph data, e.g., Facebook and Twitter with millions or even billions of nodes. What's more, the inter-connected graph structure also prevents parallel graph representation learning, which is especially critical for large-sized graph data. In addition, most of these existing graph neural networks focus on supervised learning. They encode the graph structure into representation vectors with the supervision of label information. However, for real-world graph data, manual graph labeling can be very tedious and expensive, which becomes infeasible for large-scale graphs. To overcome this challenge, some works try unsupervised learning settings instead. They optimize models with objective functions defined for capturing node proximity \cite{hamilton2017representation} or reconstructing graph structures \cite{kipf2016variational}. However, detached from supervision information, representations learned by such unsupervised approaches can hardly work well in downstream applications with specific task objectives \cite{Meng2019LATTEAO}.

    Self-supervised learning \cite{jing2020self} has recently emerged as a promising approach to overcome the dilemma of lacking available supervision. Its key idea is defining an annotation-free pretext task and generating surrogate training samples automatically to train an encoder for representation learning. In the field of computer vision, data augmentation \cite{shorten2019survey}, such as flipping or cropping, is commonly used for training sample generation, which can improve the generalization of self-supervised learning. However, due to the unordered vertexes and extensive connections in graph data, such existing techniques mentioned above cannot work anymore and new data augmentation methods for graph data specifically are needed.
    
\begin{figure}[!tbp]
     		\centering
    		\includegraphics[width=0.9\linewidth]{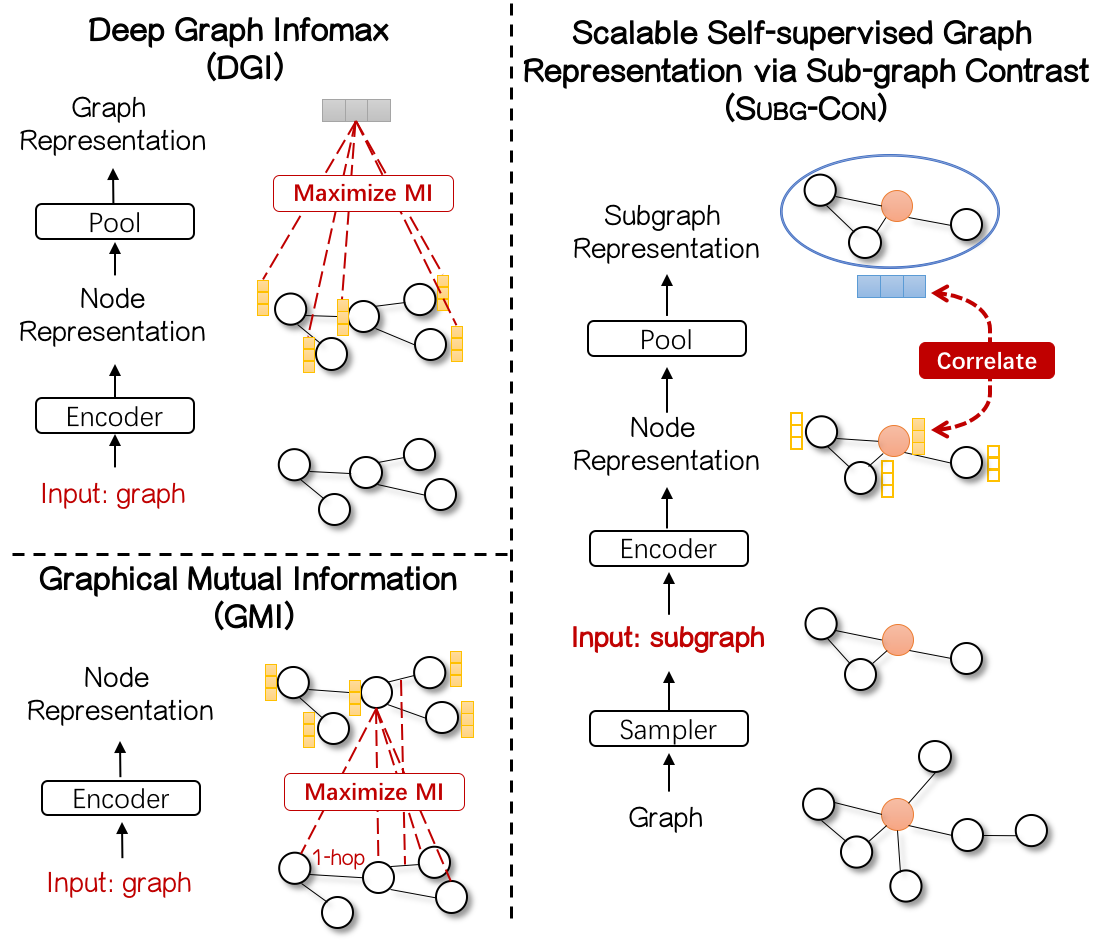}
		\vspace{-10pt}
    		\caption{An illustration of DGI (upper left), GMI (bottom left), and our proposed {\our} (right). Little colored squares denote learnt representations. Red nodes denote central nodes of context subgraphs. {\our} utilizes the strong correlation between central nodes and their context subgraphs sampled from the original graph. Note that {\our} encodes the sampled subgraphs while the other two methods take the complete graph as the input. Besides, {\our} captures structure information from regional neighborhoods instead of tending to be biased in fitting either the overall or very local (1-hop neighbor) graph structures. }
		\label{fig:task}
     		\vspace{-10pt}
 	\end{figure}

    Self-supervised graph representation learning is a new research problem, but there're still existing some prior works on this topic, e.g., Deep Graph Infomax \cite{velivckovic2018deep}  and Graphical Mutual Information \cite{peng2020graph} (even though these approaches pose themselves as unsupervised models initially). Deep Graph Infomax (DGI) \cite{velivckovic2018deep} introduces a global level pretext task to discriminate actual node representations from the corrupted ones based on the global graph representation. Graphical Mutual Information (GMI) \cite{peng2020graph} is centered about local structures by maximizing mutual information between the hidden representation of each node and the original features of its directly adjacent neighbors. As illustrated in the left of Fig. \ref{fig:task}, these works tend to be biased in fitting either the overall or very local (1-hop neighbor) graph structures in defining the mutual information based loss terms, which would harm the quality of learned representations. Besides, these self-supervised works adopt a graph neural network as the encoder and also need to take the complete graph as the input, which restricts their scalability on large-sized graphs.
    
       Intuitively, nodes and their regional neighbors are more correlated while other nodes that are very far away hardly influence them, especially in large-scale graphs. Therefore, subgraphs consisting of regional neighborhoods play a critical role to provide structure contexts for node representation learning. In this paper, we propose a novel scalable self-supervised graph representation via \underline{Sub-g}raph \underline{Con}trast, {\our}. It takes the strong correlation between central nodes and their regional subgraphs (involving both direct neighbors and other nodes that are further away) into consideration as illustrated in Fig. \ref{fig:task}. More specifically, we introduce a data augmentation strategy on graphs based on subgraph sampling firstly. The central nodes together with their closely related surrounding nodes are sampled from the original graph to compose context subgraphs. Then, these subgraphs are fed into graph neural network encoders to obtain the representations of central nodes and subgraphs after pooling. Finally, a contrastive loss is introduced in the latent space to train the encoder to distinguish the generated positive and negative samples (to be introduced later), so that nodes with different regional structures can be well-differentiated. Compared with the previous methods operating on the complete graph structure, {\our} can capture regional information in context subgraphs of smaller sizes and simpler structures with lower time and space costs. Besides, based on sampled subgraph instances, {\our} is easy to parallelize, which is critical for large-sized graph data. 
    
    Through an empirical assessment on several benchmark graph datasets with different sizes from multiple diverse fields, we demonstrate that the representations learned by our {\our} model are consistently competitive on node classification tasks, often outperforming both supervised and unsupervised strong baselines. Besides, we verify the efficiency of {\our}, both on training time and computation memory, compared with state-of-the-art self-supervised methods that work on the complete graph.

    To summarize, our major contributions include:
    \begin{itemize}
    \item We propose a novel self-supervised graph representation learning method via sub-graph contrast. It utilizes the correlation of central nodes and context subgraphs to capture regional graph structure information. 
    \item We introduce a data augmentation strategy on graphs, which aims at increasing the training samples from the existing graph using subgraph sampling for self-supervised graph representation learning. 
    \item By training with subgraphs of small sizes and simple structures, {\our} method requires lower training time and computation memory costs for graph representation learning.
    \item Based on the sampled subgraph instances, our method enables parallel graph representation learning to further improve efficiency and scalability. 
    \item Extensive experiments verify both the effectiveness and the efficiency of our work compared with prior unsupervised and supervised approaches on multiple real-world graph datasets from different domains.  
    \end{itemize}

\section{Related work}
\subsection{Graph Neural Networks}
    Graph neural networks use the graph structure as well as node features to learn node representation vectors. Existing graph neural networks follow a neighborhood aggregation strategy, which we iteratively update the representation of a node by aggregating representations of its neighboring nodes and combining with its representations \cite{xu2018powerful}. Existing graph neural networks have led to advances in multiple successful applications across different domains \cite{kipf2016semi, DBLP:journals/corr/abs-1710-10903, xu2018powerful, abu2019mixhop}. However, they usually take a complete graph as input. Thus, they can hardly be applied to large-scale graph data. What’s more, the inter-connected graph structure also prevents parallel graph representation learning, which is especially critical for large-sized graph data. To handle these issues, sampling-based methods are proposed to train GNNs based on mini-batch of nodes, which only aggregate the representations of a subset of randomly sampled nodes in the mini-batch \cite{hamilton2017inductive, chen2018fastgcn}. Although this kind of approaches reduces the computation cost in each aggregation operation, the total cost can still be large. Besides, these graph neural networks mainly focus on supervised learning and require the supervision of label information. It is intractable for them to handle unlabeled graphs, which are widely available in practically.
    
    \subsection{Unsupervised Node Representation Learning} 
    There is abundant literature in traditional unsupervised representation learning of nodes within graphs. Existing methods optimize models with random walk-based objectives \cite{perozzi2014deepwalk, grover2016node2vec, tang2015line} or reconstructing graph structures \cite{hamilton2017inductive, kipf2016variational}. The underlying intuition is to train an encoder network so that nodes that are close in the input graph are also close in the representation space. Although these methods claim to capture node proximity, they still suffer from some limitations. Most prominently, they over-emphasizing proximity similarity, making it difficult to capture the inherent graph structural information. Besides, as these encoders already enforce an inductive bias that neighboring nodes have similar representations, it is unclear whether such objectives actually provide any useful signal for training an encoder. Thus, existing methods fail to solve real-world tasks as strongly as supervised methods do. 

    \subsection{Self-supervised Learning}
    Self-supervised learning has recently emerged as a promising approach to overcome the dilemma of lacking available supervision. Its key idea is defining an annotation free pretext task and generating surrogate training samples automatically to train an encoder for representation learning. A wide variety of pretext tasks have been proposed for visual representation learning \cite{gidaris2018unsupervised, asano2019critical, chen2020simple}. However, there are a few works of literature about self-supervised methods for graph representation learning so far. Deep graph infomax \cite{velivckovic2018deep} aims to train a node encoder that maximizes mutual information between node representations and the pooled global graph representation. Graphical mutual information \cite{peng2020graph} proposes to maximize the mutual information between the hidden representation of each node and the original features of its 1-hop neighbors. These works tend to be biased in fitting either the overall or very local (1-hop neighbor) graph structures in defining the mutual information based loss terms, which would harm the quality of learned representations. Besides, these self-supervised works also need to take the complete graph as the input, which restricts their scalability on large-sized graphs.

\section{Method}
	In this section, we will present our framework in a top-down fashion. It starts with an abstract overview of our specific subgraph-based representation learning setup, followed by an exposition of subgraph sampling based data augmentation, subgraph encoding for representations, and our self-supervised pretext task for model optimization. Finally, we introduce parallel {\our} briefly.

	\subsection{Subgraph-Based Self-Supervised Representation Learning}
		Prior to going further, we first provide the preliminary concepts used in this paper. We assume a general self-supervised graph representation learning setup:  For a graph $\mathcal{G} = (\mathbf{X}, \mathbf{A})$, a set of node features are provided, $\mathbf{X} = \{ \mathbf{x}_{1},  \mathbf{x}_{2}, ..., \mathbf{x}_{N}\}$, where $N$ is the number of nodes in the graph and $\mathbf{x}_{i} \in \mathbb{R}^{F}$ represents the features of dimension $F$ for node $i$. We are also provided with relational information between these nodes in the form of an adjacency matrix, $\mathbf{A} \in \mathbb{R}^{N \times N}$. While $\mathbf{A}$ may consist of arbitrary real numbers (or even arbitrary edge features), in all our experiments we will assume the graphs to be unweighted, i.e. $\mathbf{A}(i,j) = 1$ if there exists an edge ${i} \rightarrow {j}$ in the graph and $\mathbf{A}(i,j) = 0$ otherwise.

		Traditional graph representation methods target on training an encoder $\mathcal{E} : \mathbb{R}^{N \times F} \times \mathbb{R}^{N \times N} \rightarrow \mathbb{R}^{N \times F'}$ to encode a complete graph, so that latent node representations $\mathbf{H} = \mathcal{E}(\mathbf{X}, \mathbf{A}) \in \mathbb{R}^{N \times F'}$ can be produced, where $F'$ is the dimension of latent representations. For convenience, we represent the leant representation of each node ${i}$ as $\mathbf{h}_{i}$. These representations then are generated at once and retrieved for downstream tasks, such as node classification. However, due to limited computation time and memory, it remains a great challenge for traditional methods taking the complete graph structure as the input to handle large-scale graphs. 

		To overcome the limitation of traditional methods, we propose a novel subgraph-based representation learning approach. For a central node $i$, a subgraph sampler $\mathcal{S}$, i.e., a proxy of data augmentation, is designed to extract its context subgraphs $\mathbf{X}_{i} \in \mathbb{R}^{N' \times F}$ from the original graph. The context subgraph provides regional structure information for learning the representation of node $i$. $\mathbf{X}_{i} \in \mathbb{R}^{N' \times F}$ denotes the node features of the $i$th context subgraph. $\mathbf{A}_{i}$ denotes the relational information among node $i$ and its neighbor nodes. $N'$ indicates the context subgraph size. We target at learning a encoder for context subgraphs, $\mathcal{E} : \mathbb{R}^{N' \times F} \times \mathbb{R}^{N' \times N'} \rightarrow \mathbb{R}^{N' \times F'}$, which serves for acquiring node representations within context graphs. It should be noted that different from traditional methods, the input of the encoder are context subgraphs whose sizes are much smaller than the original graph. And node representations can be retrieved based on their context subgraph structures flexibly without the complete graph. Thus, by operating on sampled subgraph instances, {\our} has prominent performance advantages in model learning scalability. Besides, it is easy to parallelize, which is critical for large-sized graph data.
		

		Here we will focus on three key points for our subgraph-based self-supervised learning method: context subgraph extraction, subgraph encoding for representations, and the self-supervised pretext task for model optimization. 
    \begin{itemize}
    \item For context subgraph extraction, the subgraph sampler $\mathcal{S}$ will serve as the proxy of data augmentation. It measures the importance scores of neighbors and samples a few closely related nodes to compose a context subgraphs which provide regional structure information for representation learning.  
    \item For subgraph encoding, we target on encoding the structures and features of context subgraphs by the encoder $\mathcal{E}$, to produce the central node representations $\mathbf{h}_{i}$. The other key consequence is summarizing the subgraph centered around node $i$ as the subgraph representation $\mathbf{s}_{i}$. 
    \item For the self-supervised pretext task, it can optimize the encoder by taking advantage of the strong correlation between central nodes and their context subgraphs so that the regional information captured from the context subgraphs embeds into the central node representations.
    \end{itemize}
	

	
	\subsection{Subgraph Sampling Based Data Augmentation}
	
	\begin{figure*}[!htbp]
     		\centering
		\vspace{-10pt}
    		\includegraphics[width=0.9\textwidth]{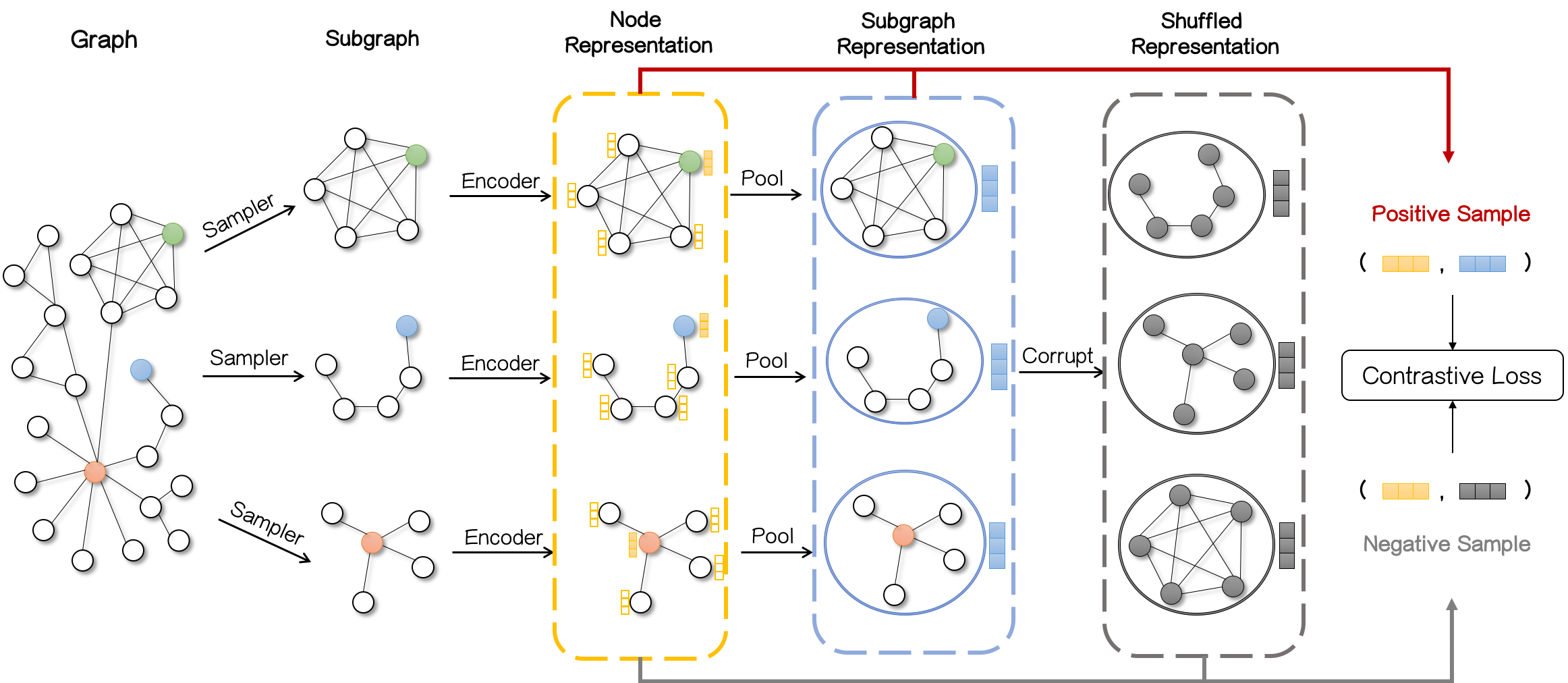}
		\vspace{-10pt}
    		\caption{ Architecture of {\our}. A series of context subgraph are sampled from the original graph and fed into the encoder to obtain the representations of central nodes and subgraphs after pooling. For a specified node, its context subgraph representation is regarded as positive sample while other subgraph representations randomly sampled are regarded as negative samples. The contrastive loss in the latent space will force the encoder to recognize positive and negative samples so that different nodes can be well-discriminated based on regional structure information. }
     		\label{fig:model}
     		\vspace{-10pt}
 	\end{figure*}
	
	To overcome the dependence of manual labels, it is important for self-supervised learning to generate surrogate training samples automatically to train an encoder for representation learning. Data augmentation is a popular technique for training sample generation in computer vision. However, due to unordered vertexes and extensive connections, it hasn't been used explicitly in graph data. For self-supervised graph representation learning, we introduce the concept of data augmentation on graph formally here.

\begin{myDef} (Data Augmentation on Graph): Given a graph $\mathcal{G} = (\mathbf{X}, \mathbf{A})$, where $\mathbf{X}$ denotes node features and  $\mathbf{A}$ denotes relations, data augmentation is a strategy to produce a series of variant graphs $\mathcal{G}' = (\mathbf{X}', \mathbf{A}')$ using various transformations on features and relations of $\mathcal{G}$ . 
\end{myDef}

	There are various transformations for graph data, such as node masking or feature corruption. In this paper, we adopt a subgraph sampling based data augmentation strategy. Because, intuitively, nodes and their regional neighborhoods are more correlated while long-distance nodes hardly influence them. This assumption is more reasonable as the size of graphs increases. Therefore, we sample a series of subgraphs including regional neighborhoods from the original graph as training data. 
    
    The most critical issue now is to sample a context subgraph, which can provide sufficient structure information for learning a high-quality representation for the central node. Here we follow the subgraph sampling based on personalized pagerank algorithm \cite{jeh2003scaling} as introduced in \cite{zhang2020graph}. Considering the importance of different neighbors varies, for a specific node $i$, the subgraph sampler $\mathcal{S}$ first measures the importance scores of other neighbor nodes by personalized pagerank algorithm. Given the relational information between all nodes in the form of an adjacency matrix, $\mathbf{A} \in \mathbb{R}^{N \times N}$, the importance score matrix $\mathbf{S}$ can be denoted as
        \begin{equation}\label{eq1}
                \mathbf{S} = \alpha \cdot (\mathbf{I} - (1 - \alpha) \cdot \bar{\mathbf{A}}) , 
        \end{equation}
        where $\mathbf{I}$ is the identity matrix and $\alpha \in [0, 1]$ is a parameter which is always set as 0.15. $\mathbf{D}$ denotes as the corresponding diagonal matrix with $\mathbf{D}(i, i) = \sum _ {j} \mathbf{A}(i, j)$ on its diagonal and $\bar{\mathbf{A}} = \mathbf{A} \mathbf{D}^{-1} $ denotes the colum-normalized adjacency matrix. $\mathbf{S}(i, :)$ is the importance scores vector for node $i$, which indicates its correlation with other nodes. 
        
        It is noted that the importance score matrix S can be precomputed before model training starts. And we implement node-wise PPR to calculate importance scores to reduce computation memory, which makes our method more suitable to work on large-scale graphs.
        
        For a specific node $i$, the subgraph sampler $\mathcal{S}$ chooses top-k important neighbors to constitute a subgraph with the score matrix $\mathbf{S}$. The index of chosen nodes can be denoted as 
        \begin{displaymath}
                idx = top\_rank(\mathbf{S}(i, :), k) , 
        \end{displaymath}
        where $top\_rank$ is the function that returns the indices of the top k values and $k$ denotes the size of context graphs. 
        
        The subgraph sampler $\mathcal{S}$ will process the original graph with the node index to obtain the context subgraph $\mathcal{G}_{i}$ of node $i$. Its adjacency matrix $\mathbf{X}_{i}$ and feature matrix $\mathbf{A}_{i}$ are denoted respectively as 
        \begin{displaymath}
            \mathbf{X}_{i} = \mathbf{X}_{idx, :}, \ \ \mathbf{A}_{i} = \mathbf{A}_{idx, idx}, 
        \end{displaymath}
        where $\cdot_{idx}$ is an indexing operation. $\mathbf{X}_{idx, :}$ is the row-wise (i.e. node-wise) indexed feature matrix. $\mathbf{A}_{idx, idx}$ is the row-wise and col-wise indexed adjacency matrix corresponding to the induced subgraph.
        
        So far, we can acquire the context subgraph $\mathcal{G}_{i} = (\mathbf{X}_{i}, \mathbf{A}_{i}) \sim \mathcal{S}(\mathbf{X}, \mathbf{A})$ for any specific node $i$. For large-sized input graphs, this procedure can support parallel computing to further improve efficiency. These context subgraphs produced by data augmentation can be decomposed into several mini-batches and fed to train {\our}. 
        
    \subsection{Encoding Subgraph For Representations}
        Given the context subgraph $\mathcal{G}_{i} = (\mathbf{X}_{i}, \mathbf{A}_{i})$ of a central node $i$, the encoder $\mathcal{E} : \mathbb{R}^{N' \times F} \times \mathbb{R}^{N' \times N'} \rightarrow \mathbb{R}^{N' \times F'}$ encodes it to obtain the latent representations matrix $\mathbf{H}_{i}$ denoted as 
        \begin{displaymath}
            \mathbf{H}_{i} = \mathcal{E}(\mathbf{X}_{i}, \mathbf{A}_{i}), 
        \end{displaymath}
        Here we adopt graph neural networks (GNN), a flexible class of node embedding architectures, as the encoder $\mathcal{E}$. Node representations are generated by aggregating information from neighbors. We study the impact of different graph neural networks in the experiments and will discuss later. The central node embedding $\mathbf{h}_{i}$ is picked from the latent representations matrix $\mathbf{H}_{i}$ 
        \begin{displaymath}
            \mathbf{h}_{i} = \mathcal{C}(\mathbf{H}_{i}),  
        \end{displaymath}
        where $\mathcal{C}$ denotes the operation picking out the central node embedding. 
        
        As mentioned before, the other key consequence is summarizing the subgraph centered around node $i$ as the context subgraph representation $\mathbf{s}_{i}$. In order to obtain the subgraph-level summary vectors, we leverage a readout function, $\mathcal{R} : \mathbb{R}^{N' \times F' } \rightarrow \mathbb{R}^{F'}$, and use it to summarize the obtained node representations into a subgraph-level representation, $\mathbf{s}_{i}$, denoted as 
        \begin{displaymath}
            \mathbf{s}_{i} = \mathcal{R}(\mathbf{H}_{i}) .   
        \end{displaymath}
        
        So far, the representations of central nodes and context subgraphs are produced, which will play a key role in the generation of positive and negative samples for self-supervised pretext tasks. 

	\subsection{Contrastive Learning via Central Node and Context Subgraph}
	The key idea of self-supervised contrastive learning is defining an annotation-free pretext task and then generating positive and negative samples. The encoder can be trained by contrasting positive and negative examples. As the prerequisite for ensuring the quality of learned representations, if the pretext task can fully inherit the rich information in graphs, we can obtain better representations to support subsequent mining tasks without additional guidance. 
         
         Intuitively, nodes are dependent on their regional neighborhoods and different nodes have different context subgraphs. This assumption is even more reasonable in large-scale graphs. At the same time, the complete structure of large-scale graphs is still hard to handle by existing node representation learning methods. Therefore, we consider the strong correlation between central nodes and their context subgraphs to design a self-supervision pretext task. The architecture of {\our} is fully summarized by Fig. \ref{fig:model}. 
         
         Our approach for learning the encoder relies on, for a specific central node, contrasting its real context subgraph with a fake one. Specifically, for the node representation, $\mathbf{h}_{i}$, that captures the regional information in the context subgraph, we regard the context subgraph representation $\mathbf{s}_{i}$ as positive sample. On the other hand, for a set of subgraph representations, we employ a function, $\mathcal{P}$, to corrupt them to generate negative samples, denoted as  
         \begin{displaymath}
             \{\widetilde{\mathbf{s}}_{1}, \widetilde{\mathbf{s}}_{2} ..., \widetilde{\mathbf{s}}_{M} \}\sim \mathcal{P}(\{\mathbf{s}_{1}, \mathbf{s}_{2}, ..., \mathbf{s}_{m}\}), 
         \end{displaymath}
         where $m$ is the size of the representation set. The corruption strategy determines the differentiation of nodes with different contexts, which is crucial for some downstream tasks, such as node classification.
		
\begin{algorithm}[htb] 
\caption{ Optimization Algorithm. } 
\label{alg:Framwork} 
\begin{algorithmic}[1] 
\Require 
A graph $\mathcal{G}$ with input feature $\mathbf{X}$ and adjacency matrix $\mathbf{A}$; Subgraph sampler $\mathcal{S}$; Encoder $\mathcal{E}$; Readout function $\mathcal{R}$; Corruption function $\mathcal{P}$.
\State Precompute importance score matrix $\mathbf{S}$ according to Eq. \ref{eq1}.
\While {not converge} 
\State Sample context subgraphs $\{(\mathbf{X}_{1}, \mathbf{A}_{1})$, $(\mathbf{X}_{2}, \mathbf{A}_{2})$, ..., $(\mathbf{X}_{m}, \mathbf{A}_{m})\}$ where $\mathbf{H}_{i} = \mathcal{S}(\mathbf{X}_{i}, \mathbf{A}_{i})$ and $m$ is the size of the subgraph set. 
\ForAll {each subgraph $(\mathbf{X}_{i}, \mathbf{A}_{i})$} 
\State Encode the subgraph to obtain latent representation matrixes $\mathbf{H}_{i} = \mathcal{E}(\mathbf{X}_{i}, \mathbf{A}_{i})$. 
\State Obtain the central node representation  $\mathbf{h}_{i} = \mathcal{C}(\mathbf{H}_{i})$. 
\State Summarize the subgraph representation through the readout function $\mathbf{s}_{i} = \mathcal{R}(\mathbf{H}_{i})$. 
\EndFor
\State Corrupt the subgraph representations to generate negative examples for the corresponding node representations $\{ \widetilde{\mathbf{s}}_{1}, \widetilde{\mathbf{s}}_{2}, ..., \widetilde{\mathbf{s}}_{M}\} = \mathcal{P}(\mathbf{s}_{1}, \mathbf{s}_{2}, ..., \mathbf{s}_{M})$.
\State Update parameters of $\mathcal{E}$ and $\mathcal{R}$ by applying gradient descent to maximize Eq. \ref{eq2}.
\EndWhile 
\end{algorithmic} 
\end{algorithm}

\begin{table*}[!htbp]
\vspace{-10pt}
\small
\center
\caption{Dataset statistics}
\label{tab:Datasets}
\begin{tabular}{ccccccccccccc}
\toprule
\  & Dataset  & Type & Nodes & Edges & Degree & Features & Classes  & Train / Val / Test \\
\midrule
\multirow{3}*{Small-scale}
 & Cora & Citation network & 2,708 & 5,429 & 4.0 & 1,433 & 7 & 0.05 / 0.18 / 0.37  \\
 & Citeseer & Citation network & 3,327 & 4,732 & 2.8 & 3,703 & 6 & 0.04 / 0.15 / 0.30  \\
 & Pubmed & Citation network & 19,717 & 44,338 & 4.5 & 500 & 3 & 0.003 / 0.03 / 0.05  \\
\midrule
\multirow{3}*{Large-scale}
 & PPI & Protein network & 56,944 & 818,716 & 28.8 & 50 & 121 & 0.79 / 0.11 / 0.10  \\
 & Flickr & Social network & 89,250 & 899,756 & 20.2 & 500 & 7 & 0.50 / 0.25 / 0.25  \\
 & Reddit & Social network & 232,965 & 11,606,919 & 99.6 & 602 & 41 & 0.66 / 0.10 / 0.24  \\
\bottomrule
\end{tabular}
\vspace{-10pt}
\end{table*}	

		As to the objective, related works use a noise-contrastive type objective with a standard binary cross-entropy loss between positive examples and negative examples \cite{velivckovic2018deep}. However, as these context subgraphs are extracted from the same original graph and overlap with each other, we suppose that it can be harmful for representation learning if positive and negative examples to be distinguished absolutely. Therefore, we use the margin triplet loss \cite{schroff2015facenet} for model optimization so that positive and negative samples can be well-discriminated to some extent and high-quality representations can be obtained. The loss is denoted as
		\begin{equation}\label{eq2}
		 	\mathcal{L} = \frac {1} {M} {\sum _ {i = 1} ^ {M} \mathbb{E}_{(\mathbf{X}, \mathbf{A})} (-\max (\sigma(\mathbf{h}_{i}\mathbf{s}_{i}) - \sigma(\mathbf{h}_{i}\widetilde{\mathbf{s}}_{i}) + \epsilon, 0) )},   
		 \end{equation}
		where $\sigma(x) = 1/(1 + \exp(-x))$ is the sigmoid function and $\epsilon$ is margin value.
		We summarize the steps of the procedure of our approach in Algorithm \ref{alg:Framwork}.

		\subsection{Parallelizability}
		Compared with existing methods that input the complete graph data, it is parallelizable to operate on context subgraph. On the one hand, subgraph extraction is easy to parallelize. Several random workers (in different threads, processes, or machines) can simultaneously explore different parts of the same graph for context subgraphs extraction. On the other hand, without the need for global computation that needs the whole graph structure, it becomes possible to encoder multiple subgraphs synchronously to obtain the representations of central nodes and subgraphs. Benefit from the parallelizability, our model can be scaled efficiently on larger-size graphs.

\begin{table*}[!htbp]
\vspace{-10pt}
\small
\center
\caption{Performance comparison with different methods on node classification. The second column illustrates the data used by each algorithm in the training phase, where $\mathbf{X}$, $\mathbf{A}$, and $\mathbf{Y}$ denotes features, adjacency matrix, and labels, respectively. \textbf{OOM}: Out of memory.} 
\label{tab:Result}
\begin{tabular}{c|c|ccc|ccc}
\toprule
 Algorithm &  Available data & Cora & Citeseer  & Pubmed & PPI & Flickr & Reddit \\ 
\midrule
 Raw features 	& $\mathbf{X}$ 			&  56.6 $\pm$ 0.4 	& 57.8 $\pm$ 0.2 	& 69.1 $\pm$ 0.2 	& 42.5 $\pm$ 0.3	& 20.3 $\pm$ 0.2	& 58.5 $\pm$ 0.1	\\
 DeepWalk 	& $\mathbf{A}$  			& 67.2 		& 43.2 		& 65.3 		& 52.9	 	& 27.9	  	& 32.4 		\\
 Unsup-GraphSAGE 		&$\mathbf{X}$, $\mathbf{A}$  	&  75.2 $\pm$ 1.5 	& 59.4 $\pm$ 0.9 	& 70.1 $\pm$ 1.4	& 46.5 $\pm$ 0.7	& 36.5 $\pm$ 1.0	& 90.8 $\pm$ 1.1	\\
 DGI 			& $\mathbf{X}$, $\mathbf{A}$  	& 82.3 $\pm$ 0.6 	& 71.8 $\pm$ 0.7 	& 76.8 $\pm$ 0.6 	& 63.8 $\pm$ 0.2	& 42.9 $\pm$ 0.1	& 94.0 $\pm$ 0.1	\\
 GMI		 	& $\mathbf{X}$, $\mathbf{A}$  	& 83.0 $\pm$ 0.3 	& 73.0 $\pm$ 0.3 	& 79.9 $\pm$ 0.2 	& 65.0 $\pm$ 0.0	& 44.5 $\pm$ 0.2	& 95.0 $\pm$ 0.0	\\
\midrule
GCN 		& $\mathbf{X}$, $\mathbf{A}$, $\mathbf{Y}$ &  81.4 $\pm$ 0.6 	& 70.3 $\pm$ 0.7	& 76.8 $\pm$ 0.6 	& 51.5 $\pm$ 0.6	& 48.7 $\pm$ 0.3	& 93.3 $\pm$ 0.1	\\
GAT		 	& $\mathbf{X}$, $\mathbf{A}$, $\mathbf{Y}$ &  83.0 $\pm$ 0.7 	& 72.5 $\pm$ 0.7 	& 79.0 $\pm$ 0.3 	& \textbf{97.3 $\pm$ 0.2}	& \textbf{OOM}	& \textbf{OOM}			\\
FastGCN		& $\mathbf{X}$, $\mathbf{A}$, $\mathbf{Y}$ &  78.0 $\pm$ 2.1 	& 63.5 $\pm$ 1.8	& 74.4 $\pm$ 0.8 	& 63.7 $\pm$ 0.6	& 48.1 $\pm$ 0.5	& 89.5 $\pm$ 1.2	\\
GraphSAGE	& $\mathbf{X}$, $\mathbf{A}$, $\mathbf{Y}$ &  79.2 $\pm$ 1.5 	& 71.2 $\pm$ 0.5	& 73.1 $\pm$ 1.4 	& 51.3 $\pm$ 3.2	& \textbf{50.1 $\pm$ 1.3}	& 92.1 $\pm$ 1.1	\\
\midrule
 {\our} 			&$\mathbf{X}$, $\mathbf{A}$  &  \textbf{83.5 $\pm$ 0.5} & \textbf{73.2 $\pm$ 0.2} & \textbf{81.0 $\pm$ 0.1} & \textbf{66.9 $\pm$ 0.2}& \textbf{48.8 $\pm$ 0.1}& \textbf{95.2 $\pm$ 0.0}\\
\bottomrule
\end{tabular}
\vspace{-10pt}
\end{table*}

\section{Experiment}
	In this section, we conduct extensive experiments to verify both the effectiveness and the efficiency of {\our} on a variety of node classification tasks on multiple real-world datasets from different domains. In each case, {\our} is used to learn node representations in a fully unsupervised manner. We compare our approach with prior unsupervised and supervised strong baselines. Besides, we analyze the design of our architecture, including the encoder architecture and the objective function. We also do experiments about the efficiency including training time and memory usage. Reducing the number of training subgraphs and parallelization are studied to further improve efficiency. Lastly, parameter sensitivity analysis helps to choose suitable parameters for our approach.

	\subsection{Datasets}
		To assess the effectiveness of the representation learned by our work, we conduct experiments on multiple real-world datasets from different domains. We choose three popular small-scale datasets wide used in related works \cite{kipf2016semi} (Cora, Citeseer and Pubmed) and three large-scale datasets to verify the scalability of our approach (PPI, Flickr, and Reddit) \cite{kipf2016semi, zeng2019graphsaint}. It includes three citation networks, two social networks, and a protein network. All datasets follow “fixed-partition” splits. Further information on the datasets can be found in Table \ref{tab:Datasets}.
        
        We set up the experiments on the following benchmark tasks: (1) classifying research papers into topics on the Cora, Citeseer and Pubmed citation networks; (2) classifying protein roles within protein-protein interaction (PPI) networks, requiring generalization to unseen networks; (3) categorizing types of images based on the descriptions and common properties of Flickr online; (4) predicting the community structure of a social network modeled with Reddit posts. 
        \textbf{Source code} The source code of {\our} is available in https://github.com/yzjiao/Subg-Con.

	\subsection{Experimental Settings}
	
	\textbf{Encoder design.} For six different datasets, we study the impact of different graph neural networks (described below) and employ distinct encoders appropriate to that setting. 
	
	For Cora, Citeseer, Pubmed and PPI, we adopt a one-layer Graph Convolutional Network (GCN) with skip connections \cite{zhang2019gresnet} as our encoder, with the following propagation rule:
		\begin{displaymath}
    			\mathcal{E}(\mathbf{X}, \mathbf{A}) = \sigma ({{\hat{\mathbf{D}}}^{-\frac{1}{2}}} {\hat{\mathbf{A}}} {{\hat{\mathbf{D}}}^{-\frac{1}{2}}} {\mathbf{X}} {\mathbf{W}} + {\hat{\mathbf{A}}} {\mathbf{W}_{skip}}) \nonumber, 
		\end{displaymath}
		where ${\hat{\mathbf{A}}} = {\mathbf{A}} + {\mathbf{I}_{N}}$ is the adjacency matrix with inserted self-loops and ${\hat{\mathbf{D}}}$ is its corresponding degree matrix. For the nonlinearity $\sigma$, we apply the perametric ReLU (PReLU) function \cite{he2015delving}. ${\mathbf{W}}$ is a learnable linear transformation applied to every node and ${\mathbf{W}_{skip}}$ is a learnable projection matrix for skip connections. 
	
	For Reddit and Flickr, we adopt a two-layer GCN model as our encoder, with the following propagation rule:
		\begin{displaymath}
    			GCN(\mathbf{X}, \mathbf{A}) = \sigma ({{\hat{\mathbf{D}}}^{-\frac{1}{2}}} {\hat{\mathbf{A}}} {{\hat{\mathbf{D}}}^{-\frac{1}{2}}} {\mathbf{X}} {\mathbf{W}}) \nonumber, 
		\end{displaymath}
		\begin{displaymath}
    			\mathcal{E}(\mathbf{X}, \mathbf{A}) = GCN(GCN(\mathbf{X}, \mathbf{A}), \mathbf{A})  \nonumber, 
		\end{displaymath}
		where the latent representations produced by the first layer of GCN are fed as the input of the second layer. 
	
	\textbf{Corruption functions.} The corruption function generates negative samples for our self-supervised task to make nodes with different contexts well-distinguished, which is important for the node classification task. For convenience of computation, given a set of context subgraph representations, our corruption function shuffles them randomly. The subgraph representation of other central nodes is regarded as the negative sample so that nodes are closely related to their context subgraphs and weakly associated with other subgraphs. For learning node representations towards other kinds of tasks, the design of appropriate corruption strategies remains an area of open research.

    \textbf{Readout functions.} For all six experimental datasets, we employ the identical readout function with a simple averaging of all the nodes’ features:
        \begin{displaymath}
                \mathcal{R}(\mathbf{H}) = \sigma (\frac {1} {N'} \sum _ {i=1} ^ {N'} \mathbf{h}_i) \nonumber, 
        \end{displaymath}
        where $\sigma$ is the logistic sigmoid nonlinearity. We assume that this simple readout is efficient for subgraphs of small sizes when we have found it to perform the best across all our experiments. 
        
\begin{table}[!htbp]
\vspace{-10pt}
\small
\center
\caption{ Studied Objective functions.}
\label{tab:LossFunction}
\begin{tabular}{ccccccccccccc}
\toprule
 Name  & Objective Function \\
\midrule
 Margin Loss & $-\max (\sigma(\mathbf{h}\mathbf{s}) - \sigma(\mathbf{h}\widetilde{\mathbf{s}}) + \epsilon, 0) $ \\
 Logistic Loss  & $ \log \sigma(\mathbf{h}\mathbf{s}) + \log \sigma(- \mathbf{h}\widetilde{\mathbf{s}}) $ \\
 BPR Loss  & $ \log \sigma(\mathbf{h}\mathbf{s} - \mathbf{h}\widetilde{\mathbf{s}}) $ \\
\bottomrule
\vspace{-10pt}
\end{tabular}
\end{table}
     
     \textbf{Objective functions.} We compare the margin loss \cite{schroff2015facenet} against other commonly used contrastive loss functions, such as logistic loss \cite{mikolov2013efficient}, and bayesian personalized ranking (BPR) loss \cite{rendle2012bpr}. Table \ref{tab:LossFunction} shows these three objective function. Their impacts will be discussed later.

	\textbf{Implementation details.}
        We implemented the baselines and {\our} using PyTorch \cite{ketkar2017introduction} and the geometric deep learning extension library \cite{Fey/Lenssen/2019}. The experiments are conducted on 8 NVIDIA TITAN Xp GPUs. {\our} is used to learn node representations in a fully unsupervised manner, followed by evaluating the node-level classification with these representations. This is performed by directly using these representations to train and test a simple linear (logistic regression) classifier. In preprocessing, we perform row normalization on Cora, Citeseer, PubMed following \cite{kipf2016semi}, and apply the processing strategy in \cite{hamilton2017inductive} on Reddit, PPI, and Flickr. Especially, for PPI, suggested by \cite{velivckovic2018deep}, we standardize the learned embeddings before feeding them into the logistic regression classifier. During training, we use Adam optimizer \cite{kingma2014adam} with an initial learning rate of 0.001 (specially, ${10}^{-5}$ on Citeseer and Reddit). The subgraph size is no more than 20 (specially, the subgraph size is 10 on Citeseer due to better performance).  The dimension of node representations is 1024. The margin value $\epsilon$ for the loss function is 0.75.
        
    \textbf{Baselines.} We choose two state-of-art self-supervised methods, DGI \cite{velivckovic2018deep} and GMI \cite{peng2020graph}, which both learn graph embeddings by leveraging mutual information maximization. Two traditional unsupervised methods, DeepWalk \cite{perozzi2014deepwalk} and unsupervised variants of GraphSAGE (abbreviated as Unsup-GraphSAGE) \cite{hamilton2017inductive} are also compared with our model. Specially, we provide results for training the logistic regression on raw input features. Besides, we report experiment results on three supervised graph neural networks, GCN \cite{kipf2016semi}, GAT \cite{DBLP:journals/corr/abs-1710-10903}, FastGCN \cite{chen2018fastgcn} and supervised GraphSAGE \cite{hamilton2017inductive}.  Notably, we reuse the metrics already reported in original papers or choose optimal hyper-parameters carefully after reproducing the code for different baselines in this paper to ensure the fairness of comparison experiments. 
    
    \textbf{Evaluation metrics.}
    For the classification task, we provide the learned embeddings across the training set to the logistic regression classifier and give the results on the test nodes \cite{peng2020graph}. Followed \cite{velivckovic2018deep}, we adopt the mean classification accuracy to evaluate the performance for three benchmark datasets (Cora, Citeseer, and Pubmed), while the micro-averaged F1 score averaged is used for the other three larger datasets (PPI, Flickr, and Reddit). 

	\subsection{Node Classification}
    The results of our comparative evaluation experiments are summarized in Table \ref{tab:Result}. The results demonstrate our strong performance can be achieved across all six datasets. Our method successfully outperforms all the competing self-supervised approaches—thus verifying the potential of methods based on graph regional structure information in the node classification domain. We further observe that all self-supervised methods are more competitive than traditional unsupervised baselines that rely on proximity-based objectives. It indicates our data augmentation strategy for self-supervised learning can make a greater contribution to models to capture high-level information in complex graphs even if these supervision signals are not frankly related to the node classification task. Besides, we particularly note that the DGI approach is competitive with the results reported for three supervised graph neural networks, even exceeding its performance on the Cora, Citeseer, Pubmed, and Reddit datasets. However, on PPI the gap is still large—we believe because our encoder is heavily dependent on node original features while available features on PPI are extremely sparse (over 40\% of the nodes having all-zero features).
	
	\subsection{Design of Architectures}
	
	\subsubsection{Design of Encoder}

\begin{table}[!htbp]
\vspace{-10pt}
\small
\center
\caption{Comparison with different graph neural network encoders.}
\label{tab:Encoders}
\begin{tabular}{c|c|c|c|cccccc}
\toprule
Dataset &  \ \ \ GCN\ \ \  & GCN+Skip  & \ \ \ GAT\ \ \  &  \ \ \ GIN\ \ \  \\ 
\midrule
Cora 	& 	82.1 &	\textbf{83.5} 	&  \textbf{83.5}  	&	83.0 \\
Citeseer	& 	72.4 	&	\textbf{73.2} 	&  73.0 	& 	73.0 \\
Pubmed 	& 	79.2 &	\textbf{81.1} 	&  80.0  	&	80.4 \\
\midrule
PPI 		& 	66.2 &	\textbf{66.9} 	&  66.8  	&	66.0 \\
Flickr	& 	\textbf{48.8}	&	48.2 	&  48.7 	& 	48.3 \\
Reddit	& 	\textbf{95.2} 	&	94.5 	&  94.9 	& 	93.9 \\
\bottomrule
\end{tabular}
\vspace{-5pt}
\end{table}

		For better architecture and performance, we conducted experiments about the design of our encoder. We choose four different graph neural networks as the encoder to learn node representation, including graph convolutional network (GCN), graph convolutional network with skip connection (GCN + Skip), graph attention network (GAT) \cite{DBLP:journals/corr/abs-1710-10903}, graph isomorphism network (GIN) \cite{xu2018powerful}. The experimental results are listed in Table \ref{tab:Encoders}. 
		
		As can be observed, GCN with skip connection can achieve the best performance on Citeseer, Pubmed, and PPI. Although GAT can be competitive on Cora, because GAT requires more training time and memory, we choose GCN with skip connection as our encoder finally. It is noted that, even if GCN is not the best choice on these three datasets, but compared in Table \ref{tab:Result}, our method with GCN as encoder still outperforms supervised GCN. For the other two larger datasets, Flickr and Reddit, 2-layer GCN is the best option. We assume higher-level information captured in the large-scale graphs can make contributions to improve the quality of the learned representations. To sum up, compared with the complete graphs with large scales and complex structures, subgraphs can be well encoded with simple graph neural networks. More expressive GNNs, such as GAT and GIN, are less suitable to handle these subgraphs.

\begin{table}[!htbp]
\vspace{-10pt}
\small
\center
\caption{Comparison with models trained with different objective functions. }
\label{tab:LossResult}
\tabcolsep0.065 in
\begin{tabular}{c|cccccccccccc}
\toprule
 \   & Cora & Citeseer & Pubmed  & PPI & Flickr & Reddit\\
\midrule
 Margin &  \textbf{83.5} &  \textbf{73.2} &  \textbf{81.0}  &  \textbf{66.9} &  \textbf{48.8} &  \textbf{95.2}\\
 Logistic  & 82.4 & 72.2 & 79.8  & 66.8 & 48.5 & 95.0\\
 BPR  & 81.7 & 72.0 & 79.9  & 66.8 & 48.6 & 94.8\\
\bottomrule
\end{tabular}
\vspace{-5pt}
\end{table}	

	\subsubsection{Effectiveness of Objective Function}

	We compare different objective functions and list the experiment results in Table \ref{tab:LossResult}. To make the comparisons fair, we tune the hyperparameters for all loss functions and report their best results. Table \ref{tab:LossResult} shows that margin loss can achieve the best performance compared with other losses. We believe, as context subgraphs extracted from the same original graph can be somewhat similar, it is not suitable to apply the loss functions that distinguish positive and negative examples absolutely. 

	\subsection{Efficiency}
	
	\subsubsection{Train with A Few Subgraphs}

 \begin{figure}[!htbp]
  \centering
  \vspace{-10pt}
  \subfigure[Small-Scale Datasets]{
    \label{fig:sub1}
    \includegraphics[width=0.46\linewidth]{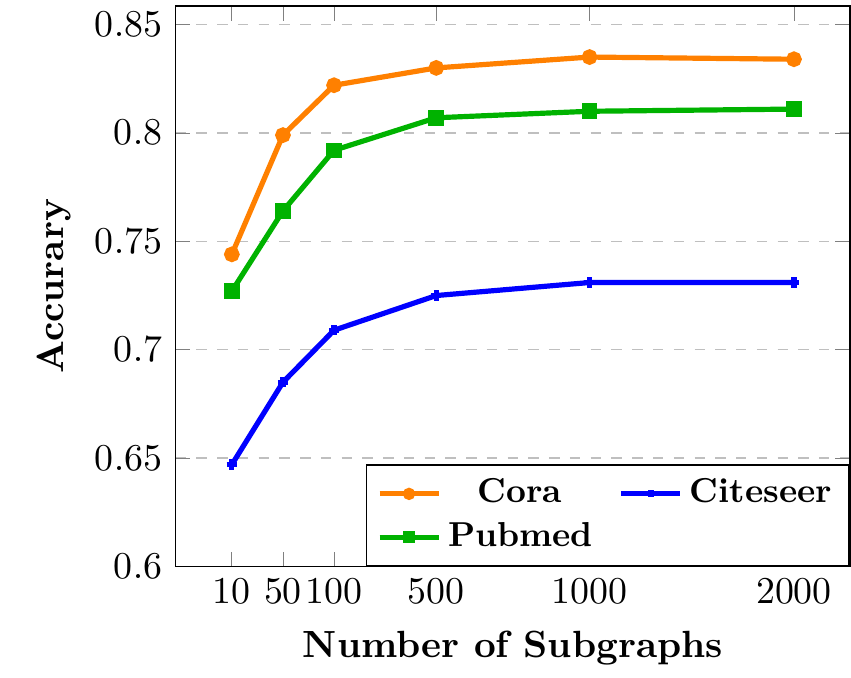}
 }
  \subfigure[Large-Scale Datasets]{
    \label{fig:sub1}
    \includegraphics[width=0.46\linewidth]{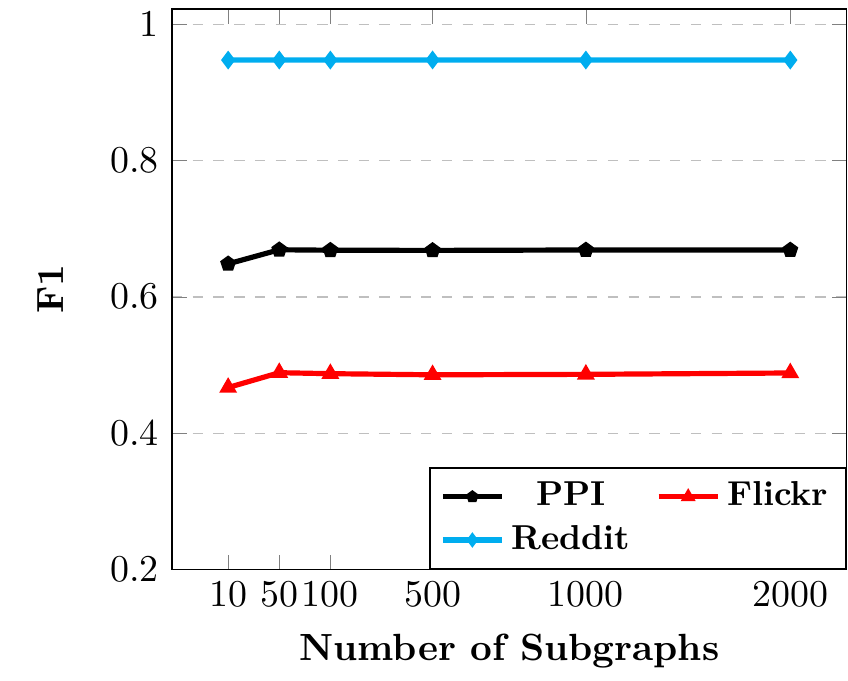}
 }
 \vspace{-5pt}
 \caption{The effectiveness of training the encoder with different numbers of sampled subgraphs}
 \label{fig:TrainSubgraphs}
 \vspace{-5pt}
\end{figure}
 
  	\begin{figure}[!tbp]
     		\centering
    		\includegraphics[width=0.7\linewidth]{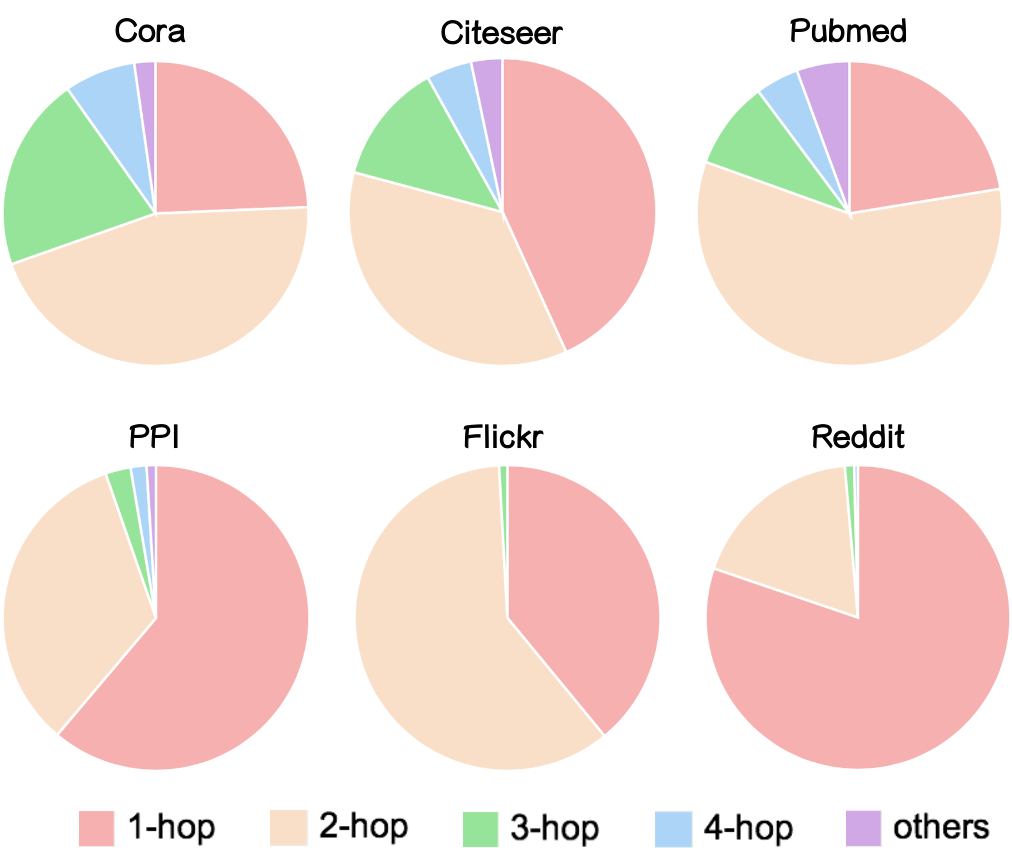}
    		\caption{Composition of context subgraphs for different datasets. The pie chart indicates the proportion of neighbors of different distances from central nodes in the context subgraphs.}
     		\label{fig:pie}
     		\vspace{-15pt}
 	\end{figure}
	
		As context subgraphs have simple and similar structures, we assume that maybe extracting all subgraphs is unnecessary for training the encoder well. Therefore, we conducted some experiments about training the encoder with a few subgraphs sampled from the graph. The effectiveness of the number of sampled subgraphs on the six datasets are showed in the Fig. \ref{fig:TrainSubgraphs}. We observed that, for Cora, Citeseer and Pubmed, about 500 subgraphs can provide sufficient information for the encoder while the other three datasets, PPI, Flickr, and Reddit, only require as few as 50 subgraphs.  We believe the sparsity of the graph leads to the difference.  The degree of nodes in Cora, Citeseer, and Pubmed is small, therefore subgraphs extracted from these datasets can be of much different shape. On the contrary, PPI, Flickr and Reddit are relatively denser and the context subgraphs likely composed of direct neighbors. Thus, the encoder can capture the structure easily. To verify our surmise, we observe the composition of subgraphs in different datasets as showed in Fig. \ref{fig:pie}. The observation guides us to train the encoder with a few subgraphs to accelerate the convergence of the loss function, which takes much less training time and computation memory.

\begin{table}[!htbp]
\small
\center
\caption{Efficiency of {\our} on three small-scale datasets. We train the encoder with 500 context subgraphs.  }
\label{tab:Efficiency1}
\begin{tabular}{ccccccccccccc}
\toprule
Dataset  & Algorithm & Training Time & Memory \\
\midrule
\multirow{3}*{Cora}
 & DGI 		& 27s 	& 3597MB	\\
 & GMI 		& 104s	& 3927MB	\\
 & {\our} 		& \textbf{14s}	& \textbf{1586MB}	\\
\midrule
\multirow{3}*{Citeseer}
 & DGI 		& 48s 	& 4867MB	\\
 & GMI 		& 410s	& 7605MB	\\
 & {\our} 		& \textbf{12s}	& \textbf{1163MB}	\\
\midrule
\multirow{3}*{Pubmed}
 & DGI 		& 104s 	& 10911MB	\\
 & GMI 		& 1012s	& 12115MB	\\
 & {\our} 		& \textbf{26s}	& \textbf{975MB}	\\
\bottomrule
\end{tabular}
\vspace{-5pt}
\end{table}

\begin{table}[!htbp]
\vspace{-10pt}
\small
\center
\caption{Efficiency of {\our} on three large-scale datasets. We train the encoder with 50 context subgraphs.  }
\label{tab:Efficiency2}
\begin{tabular}{ccccccccccccc}
\toprule
Dataset  & Algorithm & Training Time & Memory \\
\midrule
\multirow{3}*{PPI}
 & DGI 		& 44s	& 10171MB	\\
 & GMI 		& 561s	& 12101MB	\\
 & {\our} 		& \textbf{3s}		& \textbf{1349MB}	\\
\midrule
\multirow{3}*{Flickr}
 & DGI 		& 518s	& 5028MB	\\
 & GMI 		& 1247s	& 9768MB	\\
 & {\our} 		& \textbf{12s}	& \textbf{1903MB}	\\
\midrule
\multirow{3}*{Reddit}
 & DGI 		& 4071s	& 8517MB	\\
 & GMI 		& 9847s	& 12098MB	\\
 & {\our} 		& \textbf{25s}	& \textbf{3805MB}	\\
\bottomrule
\end{tabular}
\vspace{-10pt}
\end{table}

	\subsubsection{Training time and memory cost}
	
		In Table \ref{tab:Efficiency1} and Table \ref{tab:Efficiency2}, we summarize the performance on the state-of-the-arts self-supervised methods over their training time and memory usage relative to that of our method on all the six datasets. The training time refers to the time for training the encoder (exclude validation). The memory refers to total memory costs of model parameters and all hidden representations of a batch. The two self-supervised baselines apply GCN as their encoders on Cora, Citeseer, and Pubmed, which cannot be trained on large-scale graphs due to excessive memory requirements. For other larger graphs, they choose GraphSAGE, a fast sampling-based graph neural network, for node representation learning.  We use an early stopping strategy on the observed results on the validation set, with a patience of 20 epochs (specially, 150 epochs for Pubmed). According to the findings in the previous subsection, 500 subgraphs randomly sampled are used to train the encoder for three small-scale datasets in Table \ref{tab:Efficiency1} while 50 subgraphs are used for three larger datasets in Table \ref{tab:Efficiency2}. We can clearly found ours methods can be trained much faster with much less computation memory than these baselines on all the datasets. In particular, our advantage of efficiency can be more prominent on large-scale graphs, especially on Reddit. We believe that compared to the whole graph structure, subgraphs of much small size can speedup encoder training. Besides, training with a few subgraphs can further reduce training time and memory usage. 

\begin{figure}[!htbp]
  \centering
  \vspace{-10pt}
  \subfigure[Training Time]{
    \label{fig:para1}
    \includegraphics[width=0.45\linewidth]{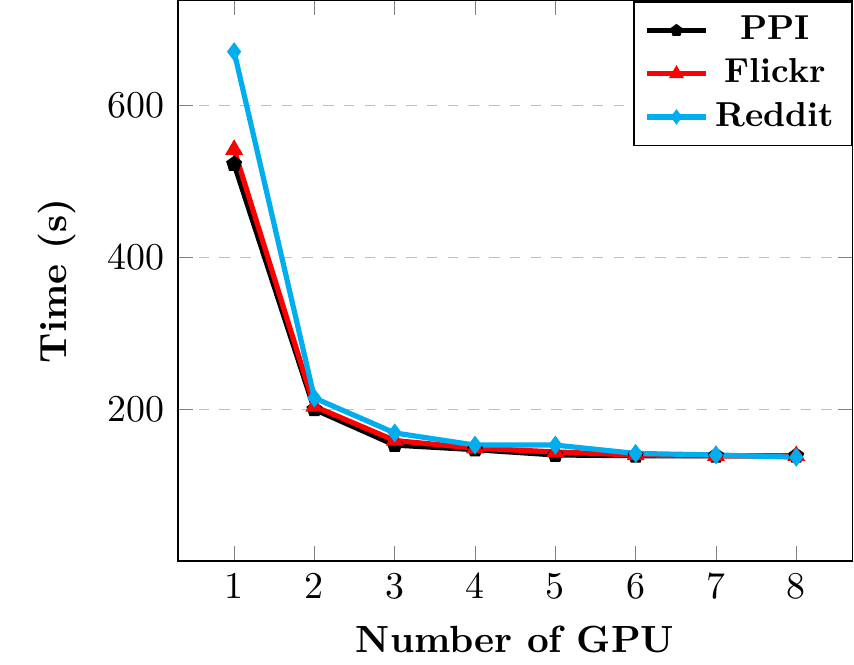}
 }
  \subfigure[Performance]{
    \label{fig:para2}
    \includegraphics[width=0.45\linewidth]{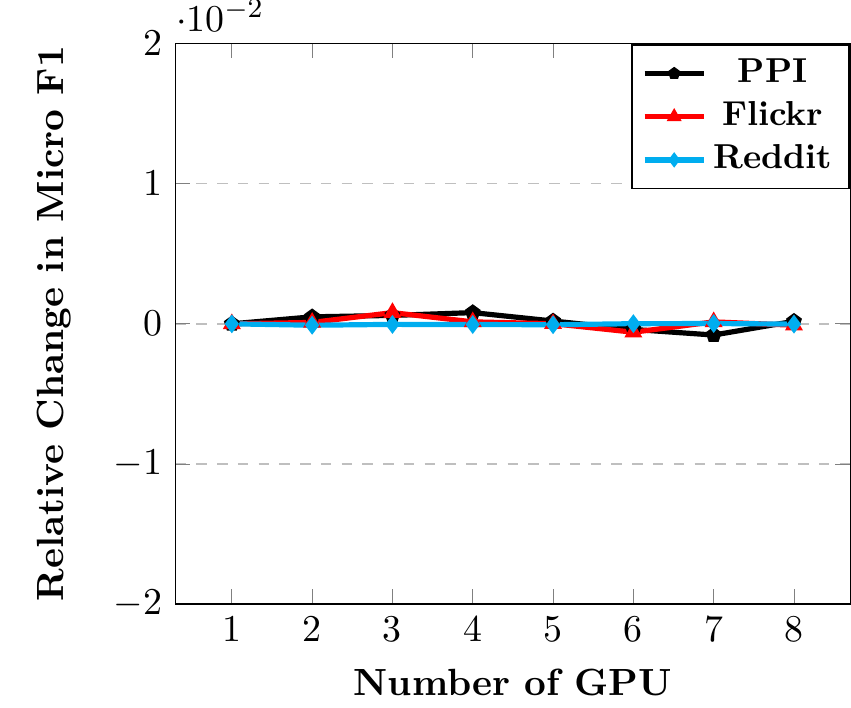}
 }
 \caption{Effects of parallelizing. }
 \label{fig:para}
\vspace{-10pt}
\end{figure}
       
       	\subsubsection{Parallel Computation}
	
	For complex realistic application scenarios, in the case when training with a small number of subgraphs doesn't work, {\our} can be run efficiently in parallel. We set the number of subgraphs as 20000 and set training epoch as 400, and run experiments using multiple GPUs on three large-scale datasets, PPI, Flickr and Reddit. Fig. \ref{fig:para} presents the effects of parallelizing. It shows the speed of processing these three data sets can be accelerated by increasing the number of GPUs (Fig. \ref{fig:para1}). It also shows that there is no loss of predictive performance relative to the running our model serially (Fig. \ref{fig:para2}). It has been demonstrated that this technique is highly scalable.

    \subsection{Subgraph size analysis}
    
\begin{figure}[!tbp]
     		\centering
		\setlength{\abovecaptionskip}{-0.1cm}   
		\includegraphics[width=0.65\linewidth]{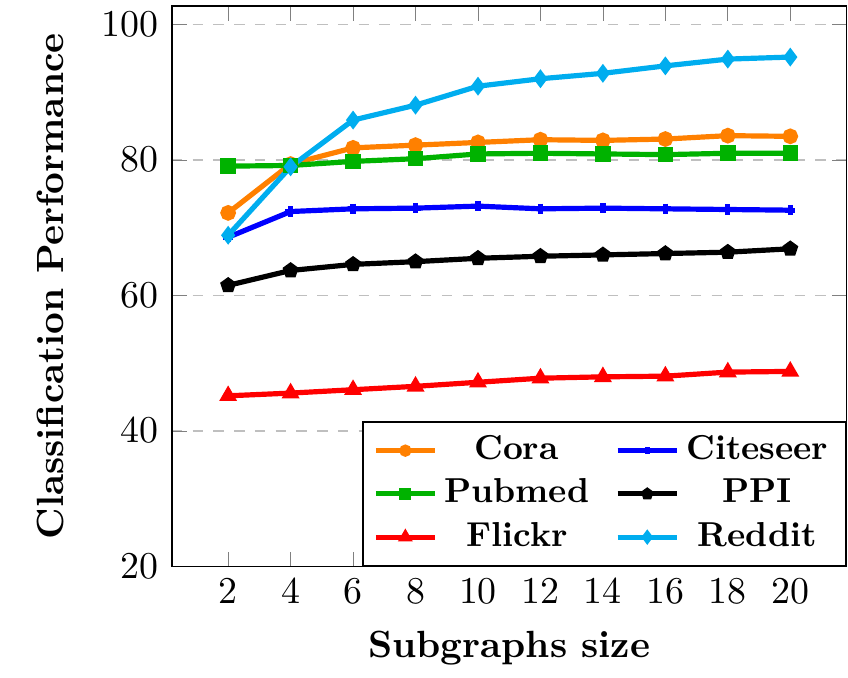}
    		\caption{Subgraph size analysis.}
     		\label{fig:size}
     		\vspace{-0.5cm}
 	\end{figure}

    Now we examine the influence of the size of context subgraphs in our framework on the six datasets. We adjust the subgraph size from 2 to 20 (including the central node) and evaluated the results as showed in Fig. \ref{fig:size}. We observe that our model can achieve better performance with context subgraphs of a larger size in general. We believe that is because more regional structure information makes a contribution to high-quality latent representations. Due to the limited computation memory, we set the subgraph size as 20. However, there is an exception. As the size of subgraphs increases, the performance on Citeseer becomes better first, reaches the peak when the size is 10, and then goes down. We consider, due to the sparsity of Citeseer, the subgraphs composed of 10 nodes have sufficient context information. Larger subgraphs with complex structures will bring about more noise and deteriorate the process of representations learning. Thus, we set the subgraph size as 10 for Citeseer. 
    It is noted that using very small subgraphs causes different impacts on different datasets. Specifically, when we train the encoder with subgraphs containing only two nodes (a central node and a closest related neighbor), the performance degrades on all the datasets. Especially, the decrease of F1 score on Reddit is up to 20 points. It indicates that Reddit is large in scale and complex in structure, therefore, a few neighbors are insufficient to be a proxy of relatively informative context. We should take it into consideration for model design.
	
\section{Conclusion}
	In this paper, we propose a novel scalable self-supervised graph representation via sub-graph contrast, {\our}. It utilizes the strong correlation between central nodes and their regional subgraphs for model optimization. Based on sampled subgraph instances, {\our} has prominent performance advantages in  weaker supervision requirements, model learning scalability, and parallelization. 
	
	Through an empirical assessment on multiple benchmark datasets, we demonstrate that the effectiveness and efficiency of {\our} compared with both supervised and unsupervised strong baselines. In particular, it shows that the encoder can be trained well on the current popular graph datasets with a little regional information. It indicates that existing methods may still lack the ability to capture higher-order information, or our existing graph dataset only requires low-order information to get good performance. We hope that our work can inspire more research on graph structure to explore the above problems. 
	
\section{Acknowledgement}
	We appreciate the comments from anonymous reviewers which will help further improve our work. This work is funded in part by the National Natural Science Foundation of China Projects No. U1636207 and No. U1936213. This work is also partially supported by NSF through grant IIS-1763365 and by FSU. This work is funded by Ant Financial through the Ant Financial Science Funds for Security Research.
	
\bibliographystyle{./IEEEtran}
\bibliography{IEEEexample}

\end{document}